\pdfoutput=1
\documentclass[11pt,a4paper]{article}
\usepackage[hyperref]{eacl2021}
\usepackage{times}
\usepackage{latexsym}
\usepackage{tabularx}
\usepackage{multirow}
\usepackage{booktabs}
\usepackage{xstring}
\usepackage{wrapfig}
\usepackage{graphics}
\usepackage{soul}

\usepackage{microtype}

\aclfinalcopy 
\usepackage{arydshln}
\usepackage{float}

\newcommand{\camelbert}{CAMeLBERT}
\newcommand{\camelbertda}{CAMeLBERT-DA}
\newcommand{\camelbertca}{CAMeLBERT-CA}
\newcommand{\camelbertmsa}{CAMeLBERT-MSA}
\newcommand{\camelbertmix}{CAMeLBERT-Mix}
\newcommand{\camelbertstar}{CAMeLBERT-Star}
\newcommand{\hide}[1]{}

\title{The Interplay of Variant, Size, and Task Type\\ in Arabic Pre-trained Language Models}

\author{Go Inoue, Bashar Alhafni, Nurpeiis Baimukan, Houda Bouamor,\textsuperscript{\textdagger} Nizar Habash \\
  Computational Approaches to Modeling Language (CAMeL) Lab\\
  New York University Abu Dhabi \\
  \textsuperscript{\textdagger}Carnegie Mellon University in Qatar\\
  \texttt{\{go.inoue,alhafni,nurpeiis,nizar.habash\}@nyu.edu}\\
  \texttt{hbouamor@qatar.cmu.edu}
  }

\date{}

\begin{document}
\maketitle
\begin{abstract}
In this paper, we explore the effects of language variants, data sizes, and fine-tuning task types in Arabic pre-trained language models. 
To do so, we build three pre-trained language models across three variants of Arabic: Modern Standard Arabic (MSA), dialectal Arabic, and classical Arabic, in addition to a fourth language model which is pre-trained on a mix of the three.
We also examine the importance of pre-training data size by building additional models that are pre-trained on a scaled-down set of the MSA variant.
We compare our different models to each other, as well as to 
eight publicly available models by fine-tuning them on five NLP tasks spanning 12 datasets.
Our results suggest that the variant proximity of pre-training data to fine-tuning data is more important than the pre-training data size.
We exploit this insight in defining an optimized system selection model for the studied tasks. 

\end{abstract}

\section{Introduction}
Pre-trained language models such as BERT~\cite{devlin2019bert} and RoBERTa~\cite{liu2019roberta} have shown significant success in a wide range of natural language processing (NLP) tasks in various languages.
Arabic has benefited from extensive efforts in building dedicated pre-trained language models, achieving state-of-the-art results in a number of NLP tasks, across both Modern Standard Arabic (MSA) and Dialectal Arabic (DA)~\cite{antoun-etal-2020-arabert,abdulmageed2020arbert}.

However, it is hard to compare these models to understand what contributes to their performances because of their different design decisions and hyperparameters, such as data size, language variant, tokenization, vocabulary size, number of training steps, and so forth.
Practically, one may empirically choose the best performing pre-trained model by fine-tuning it on a particular task; however, it is still unclear why a particular model is performing better than another and what design choices are contributing to its performance. 

To answer this question, we pre-trained various language models as part of a controlled experiment where we vary pre-training data sizes and language variants while keeping other hyperparameters constant throughout pre-training.
We started by scaling down MSA pre-training data size to measure its impact on performance in fine-tuning tasks.
We then pre-trained three different variants of Arabic: MSA, DA, and classical Arabic (CA), as well as a mix of these three variants.

We evaluate our models along with eight other recent Arabic pre-trained models across five different tasks covering all the language variants we study, namely, named entity recognition (NER), part-of-speech (POS) tagging, sentiment analysis, dialect identification, and poetry classification, spanning 12 datasets.

Our contributions can be summarized as follows:

\begin{itemize}
\setlength{\itemsep}{0pt}

\item We create and release eight Arabic pre-trained models, which we name \camelbert, with different design decisions, including one (\camelbertmix) that is trained on the largest dataset to date.\footnote{Our pre-trained models are available at \url{https://huggingface.co/CAMeL-Lab}, and the fine-tuning code and models are available at \url{https://github.com/CAMeL-Lab/CAMeLBERT}.} 
    
\item We investigate the interplay of data size, language variant, and fine-tuning task type through controlled experimentation.
Our results show that variant proximity of pre-training data and task data is more important than pre-training data size. 

\item We exploit this insight in defining an optimized system selection model.

\end{itemize}


\begin{table*}[t]
\setlength{\tabcolsep}{2pt}
\centering

\begin{tabular}{clcrrrrcr}
\toprule
\textbf{Ref} & \textbf{Model} & \textbf{Variants} & \multicolumn{1}{c}{\textbf{Size}} & \textbf{\#Word} & \multicolumn{1}{c}{\textbf{Tokens}} & \textbf{Vocab} & \textbf{\#Steps} \\ \hline
& BERT~\cite{devlin2019bert} & - & - & 3.3B & WP & 30k & 1M \\ \hline
$X_1$ & mBERT~\cite{devlin2019bert} & MSA & - & - & WP & 120k & -  \\
$X_2$ & AraBERTv0.1~\cite{antoun-etal-2020-arabert} & MSA & 24GB & - & SP & 60k & 1.25M \\
$X_3$ & AraBERTv0.2~\cite{antoun-etal-2020-arabert} & MSA & 77GB & 8.6B & WP & 60k & 3M \\
$X_4$ & ArabicBERT~\cite{safaya2020kuisail} & MSA & 95GB & 8.2B & WP & 32k & 4M \\
$X_5$ & Multi-dialect-Arabic-BERT  & MSA/DA & - & - & WP & 32k & - \\
   & \multicolumn{1}{r}{\cite{talafha2020multidialect}}  &   &   &   &   &  &   \\
$X_6$ & GigaBERTv4~\cite{lan-etal-2020-empirical} & MSA & - & 10.4B & WP & 50k & 1.48M \\
$X_7$ & MARBERT~\cite{abdulmageed2020arbert} & MSA/DA & 128GB & 15.6B & WP & 100K & 17M \\
$X_8$ & ARBERT~\cite{abdulmageed2020arbert} & MSA & 61GB & 6.5B & WP & 100K & 8M \\ \hline
& \camelbertmsa & MSA & 107GB & 12.6B & WP & 30k & 1M \\
& \camelbertda & DA & 54GB & 5.8B & WP & 30k & 1M \\
& \camelbertca & CA & 6GB & 847M & WP & 30k & 1M \\
& \camelbertmix & MSA/DA/CA & 167GB & 17.3B & WP & 30k & 1M \\
\bottomrule
\end{tabular}

\caption{\label{table:model_comparison}
Configurations of existing models and {\camelbert} models.
Ref is a model identifier used in Table~\ref{table:all_models}.
WP is WordPiece and SP is SentencePiece.
}
\end{table*}

\section{Related Work}
\label{sec:related}
There have been several research efforts on Arabic pre-trained models achieving state-of-the-art results in a number of Arabic NLP tasks.
One of the earliest efforts includes AraBERT~\cite{antoun-etal-2020-arabert}, where they pre-trained a monolingual BERT model using 24GB of Arabic text in the news domain.
\newcite{safaya2020kuisail} pre-trained ArabicBERT using 95GB of text mainly from the Arabic portion of the OSCAR corpus.
Based on ArabicBERT, \newcite{talafha2020multidialect} further pre-trained their model using 10 million tweets, which included dialectal data.
\newcite{lan-etal-2020-empirical} released several English-Arabic bilingual models dubbed GigaBERTs, where they studied the effectiveness of cross-lingual transfer learning and code-switched pre-training using Wikipedia, Gigaword, and the OSCAR corpus.
Most recently, \newcite{abdulmageed2020arbert} developed two models, ARBERT and MARBERT, pre-trained on a large collection of datasets in MSA and DA.
They reported new state-of-the-art results on the majority of the datasets in their fine-tuning benchmark.

Moreover, there have been various studies explaining why pre-trained language models perform well on downstream tasks either in monolingual \cite{hewitt-manning-2019-structural, jawahar-etal-2019-bert, liu-etal-2019-linguistic, tenney-etal-2019-bert, tenney2018what} or multilingual settings \cite{wu-dredze-2019-beto, chi-etal-2020-finding, kulmizev-etal-2020-neural, vulic-etal-2020-probing}.
Most of these efforts leveraged probing techniques to explore the linguistic knowledge that is captured by pre-trained language models such as morphosyntactic and semantic knowledge.
More recently, there have been additional efforts investigating the effects of pre-training data size and tokenization on the performance of pre-trained language models.
\newcite{zhang2020need} showed that pre-training RoBERTa requires 10M to 100M words to learn representations that reliably encode most syntactic and semantic features.
However, a much larger quantity of data is needed for the model to perform well on typical downstream NLU tasks.
\newcite{rust2020good} empirically compared multilingual pre-trained language models to their monolingual counterparts on a set of nine typologically diverse languages.
They showed that while the pre-training data size is an important factor, the designated tokenizer of each monolingual model plays an equally important role in the downstream performance.

In this work, we primarily focus on understanding the behavior of pre-trained models against variables such as data sizes and language variants.
We compare against eight existing models.
We find that AraBERTv02 ($X_3$) is the best on average and it wins or ties for a win in six out of 12 subtasks.
Our {\camelbertstar} model is second overall on average, and it wins or ties for a win in four out of 12 subtasks.
Interestingly, these systems are complementary in their performance and between the two, they win or tie for a win in nine out of 12 subtasks.

\section{Pre-training {\camelbert}}
\label{sec:pre-training}
We describe the datasets and the procedure we use to pre-train our models.
We use the original implementation released by Google for pre-training.\footnote{\url{https://github.com/google-research/bert}}

\subsection{Data}
\paragraph{MSA Training Data}
For MSA, we use the Arabic Gigaword Fifth Edition \cite{Parker:2011:arabic}, Abu El-Khair Corpus \cite{elkhair201615}, OSIAN corpus \cite{zeroual-etal-2019-osian}, Arabic Wikipedia,\footnote{\url{https://archive.org/details/arwiki-20190201}} and the unshuffled version of the Arabic OSCAR corpus \cite{ortiz-suarez-etal-2020-monolingual}.

\paragraph{DA Training Data} For DA, we collect a range of dialectal corpora: LDC97T19-CALLHOME Transcripts \cite{Gadalla:1997:callhome}; LDC2002T38-CALLHOME Supplement Transcripts \cite{ldc_callhome_ara_transcr_suppl_2002_t38}; LDC2005S08-Babylon Levantine Arabic Transcripts \cite{ldc_bab_arabic_2005_s08}; LDC2005S14-CTS Levantine Arabic Transcripts \cite{ldc_cts_levarabic_td_2005_s14}; LDC2006T07-Levantine Arabic Transcripts \cite{ldc_cts_lev_ara_td5_t_2006_t07}; LDC2006T15-Gulf Arabic Transcripts \cite{ldc_arb_gulf_cttr_2006_t15}; LDC2006T16-Iraqi Arabic Transcripts \cite{ldc_arb_iraq_cttr_2006_t16}; LDC2007T01-Levantine Arabic Transcripts \cite{ldc_arb_lev_cttr_2007_t01}; LDC2007T04-Fisher Levantine Arabic Transcripts \cite{ldc_la_cts_ann_2007_t04}; Arabic Online Commentary Dataset (AOC) \cite{Zaidan:2011:arabic};  LDC2012T09-English/Arabic Parallel text \cite{ldc_ara_dialect_eng_para_2012_t09}; Arabic Multi Dialect Text Corpora \cite{Almeman:2013:automatic}; A Multidialectal Parallel Corpus of Arabic \cite{Bouamor:2014:multidialectal}; Multi-Dialect, Multi-Genre Corpus of Informal Written Arabic \cite{Cotterell:2014:multi-dialect}; YouDACC \cite{Salama:2014:youdacc}; PADIC \cite{Meftouh:2015:machine}; Curras \cite{Jarrar:2016:curras}; WERd \cite{DBLP:conf/asru/AliN0R17}; LDC2017T07-BOLT Egyptian SMS \cite{ldc_bolt_sms_chat_ara_src_transliteration_2017_t07}; Shami \cite{kwaik2018shami}; SUAR \cite{Al-Twairesh:2018:suar}; Arap-Tweet \cite{Zaghouani:2018:araptweet}; Gumar \cite{khalifa2018gumar}; MADAR \cite{Bouamor:2018:madar}; Habibi \cite{elhaj2020habibi}; NADI \cite{mageed-etal-2020-nadi}; and QADI \cite{abdelali2020qadi}.

\paragraph{CA Training Data}
For CA, we use the OpenITI corpus (v1.2)~\cite{lorenz_nigst_2020_3891466}.

\subsection{Pre-processing}
After extracting the raw text from each corpus, we apply the following pre-processing.
We first remove invalid characters and normalize white spaces using the utilities provided by the original BERT implementation.
We also remove lines without any Arabic characters.
We then remove diacritics and kashida using CAMeL~Tools~\cite{obeid-etal-2020-camel}.
Finally, we split each line into sentences with a heuristic-based sentence segmenter.

\subsection{Preparing Data for BERT Pre-training}
We follow the original English BERT model's hyperparameters for pre-training.
We train a WordPiece~\cite{schuster2012} tokenizer on the entire dataset (167 GB text) with a vocabulary size of 30,000 using Hugging~Face's tokenizers.\footnote{\url{https://github.com/huggingface/tokenizers}}
We do not lowercase letters nor strip accents. 
We use whole word masking and a duplicate factor of 10.
We set maximum predictions per sequence to 20 for the datasets with a maximum sequence length of 128 tokens and 80 for the datasets with a maximum sequence length of 512 tokens.

\subsection{Pre-training Procedure}
We use a Google Cloud TPU (v3-8) for model pre-training.
We use a learning rate of 1e-4 with a warmup over the first 10,000 steps.
We pre-trained our models with a batch size of 1,024 sequences with a maximum sequence length of 128 tokens for the first 900,000 steps.
We then continued pre-training with a batch size of 256 sequences with a maximum sequence length of 512 tokens for another 100,000 steps.
In total, we pre-trained our models for one million steps.
Pre-training one model took approximately 4.5 days.

\begin{table*}[!htbp]
\setlength{\tabcolsep}{2pt}
\centering
\begin{tabular}{ll|rrrc|cc}
\toprule
\textbf{Task} & \textbf{Dataset/Subtask} & \textbf{\#Label} & \textbf{\#Train} & \textbf{\#Test} & \textbf{Unit} & \textbf{Variant} & \textbf{\%MSA} \\ \hline
NER & ANERcorp \cite{Benajiba:2007:ANERcorp} & 9 & 125,102 & 25,008 & Token & MSA & 83.6 \\
\hline
\multirow{3}{*}{POS} & PATB (MSA) \cite{Maamouri:2004:patb} & 32 & 503,015 & 63,172 & Token & MSA & 85.1 \\
& ARZTB (EGY) \cite{Maamouri:2012:arz} & 33 & 133,751 & 20,464  & Token & DA & 21.5 \\
& Gumar (GLF) \cite{khalifa2018gumar} & 35 & 162,031 & 20,100  & Token & DA & 30.0 \\
\hline
\multirow{3}{*}{SA} & ASTD \cite{nabil-etal-2015-astd} & 3 & 23,327 & 663 & Sent & MSA & 56.9 \\
& ArSAS \cite{ELMADANY18.22} & 3 & 23,327 & 3,705 & Sent & MSA & 60.5 \\
& SemEval \cite{Rosenthal:2017:semeval-2017} & 3 & 23,327 & 6,100 & Sent & MSA & 77.7 \\
\hline
\multirow{4}{*}{DID} & MADAR-26 \cite{salameh-etal-2018-fine} & 26 & 41,600 & 5,200  & Sent & DA & 14.1 \\
& MADAR-6 \cite{salameh-etal-2018-fine} & 6 & 54,000 & 12,000  & Sent & DA & 17.2 \\
& MADAR-Twitter-5 \cite{bouamor2019madar} & 21 & 39,836 & 9,116  & Grp & MSA & 92.3 \\
& NADI \cite{mageed-etal-2020-nadi} & 21 & 21,000 & 5,000  & Sent & DA & 38.3 \\
\hline
Poetry & APCD \cite{yousef2019learning} & 23 & 1,391,541 & 173,963  & Sent & CA & 71.3 \\
\bottomrule
\end{tabular}
\caption{\label{dataset}
Statistics of our fine-tuning datasets.
Unit refers to a unit we use to calculate the number of examples in Train and Test. 
For MADAR-Twitter-5, we use a group of five tweets as a unit (Grp).
A language variant is determined based on dataset design and the estimated proportion of MSA sentences in the dataset.
}
\end{table*}

\section{Fine-tuning Tasks}
\label{sec:fine-tuning}
We evaluate our pre-trained language models on five NLP tasks: NER, POS tagging, sentiment analysis, dialect identification, and poetry classification.
Specifically, we fine-tune and evaluate the models using 12 datasets (corresponding to 12 subtasks).
We used Hugging~Face's transformers~\cite{wolf2020huggingfaces} to fine-tune our {\camelbert} models.\footnote{We used transformers v3.1.0 along with PyTorch v1.5.1}
The fine-tuning was done by adding a fully connected linear layer to the last hidden state.

\paragraph{Tasks and Variants}
We selected the fine-tuning datasets and subtasks to represent multiple variants of Arabic by design.
For some of the datasets, the variant is readily known.
However, other datasets contain a lot of social media text where the dominant variant of Arabic is unknown.
Therefore, we estimate the proportion of MSA sentences in each dataset by identifying whether the text is MSA or DA using the Corpus 6 dialect identification model in \newcite{salameh-etal-2018-fine} as implemented in CAMeL Tools~\cite{obeid-etal-2020-camel}.
This technique does not model CA.
Of course, none of the datasets was purely MSA or DA; however, based on known dataset variants, we observe that having about 40\% or fewer MSA labels strongly suggests that the dataset is dialectal (or a strong dialectal mix).
 
Table~\ref{dataset} presents the number of labels, size, unit, variant, and MSA percentage for the datasets used in the subtasks.

\subsection{Named Entity Recognition}\label{ner}
\paragraph{Dataset}
We fine-tuned our models on the publicly available Arabic NER Dataset ANERcorp ($\sim$150K words) \cite{Benajiba:2007:ANERcorp} which is in MSA and we followed the splits defined by~\newcite{obeid-etal-2020-camel}.
We also kept the same IOB (inside, outside, beginning) tagging format defined in the dataset covering four classes: Location (LOC), Miscellaneous (MISC), Organization (ORG), and Person (PERS).

\paragraph{Experimental Setup}
During fine-tuning, we used the representation of the first sub-token as an input to the linear layer.
All models were fine-tuned on a single GPU for 3 epochs with a learning rate of 5e-5, batch size of 32, and a maximum sequence length of 512.
Since ANERcorp does not have a dev set, we used the last checkpoint after the fine-tuning is done to report results on the test set using the micro $F_1$ score. 

\subsection{Part-of-Speech Tagging}
\paragraph{Dataset} We fine-tuned our models on three different POS tagging datasets: (1) the Penn Arabic Treebank (PATB)~\cite{Maamouri:2004:patb} which is in MSA and includes 32 POS tags; (2) the Egyptian Arabic Treebank (ARZATB)~\cite{Maamouri:2012:arz} which is in Egyptian (EGY) and includes 33 POS tags; and (3) the GUMAR corpus~\cite{khalifa2018gumar} which is in Gulf (GLF) and includes 35 POS tags. 

\paragraph{Experimental Setup}
Similar to NER, we used the representation of the first sub-token as an input to the linear layer.
Our models were fine-tuned on a single GPU for 10 epochs with a learning rate of 5e-5, batch size of 32, and a maximum sequence length of 512.
We used the same hyperparameters for the fine-tuning across the three POS tagging datasets.
After the fine-tuning, we used the best checkpoints based on the dev sets to report results on the test sets using the accuracy score.

\subsection{Sentiment Analysis}
\paragraph{Dataset}
We used a combination of sentiment analysis datasets to fine-tune our models.
The datasets are: (1) the Arabic Speech-Act and Sentiment Corpus of Tweets (ArSAS) \cite{ELMADANY18.22}; (2) the Arabic Sentiment Tweets Dataset (ASTD) \cite{nabil-etal-2015-astd}; (3) SemEval-2017 task 4-A benchmark dataset \cite{Rosenthal:2017:semeval-2017}; and (4) the Multi-Topic Corpus for Target-based Sentiment Analysis in  Arabic Levantine Tweets (ArSenTD-Lev) \cite{baly2019arsentdlev}. 
We combined and preprocessed the datasets in a similar way to what was done by \newcite{AbuFarha:2019:mazajak} and \newcite{obeid-etal-2020-camel}. 
That is, we removed diacritics, URLs, and Twitter usernames from all the tweets.

\paragraph{Experimental Setup}
Our models were fine-tuned on ArSenTD-Lev and the train splits from SemEval, ASTD, and ArSAS (23,327 tweets) on a single GPU for 3 epochs with a learning rate of 3e-5, batch size of 32, and a maximum sequence length of 128.
After the fine-tuning, we used the best checkpoint based on a single dev set from SemEval, ASTD, and ArSAS to report results on the test sets.
We used the $F_{1}^{PN}$ score which was defined in the SemEval-2017 task 4-A; $F_{1}^{PN}$ is the macro $F_1$ score over the positive and negative classes only while neglecting the neutral class. 

\subsection{Dialect Identification}
\paragraph{Dataset}
We fine-tuned our models on four different dialect identification datasets: (1) MADAR Corpus 26 which includes 26 labels; (2) MADAR Corpus 6 which includes six labels; (3) MADAR Twitter Corpus~\cite{Bouamor:2018:madar,salameh-etal-2018-fine,bouamor2019madar} which includes 21 labels; and (4) NADI Country-level~\cite{mageed-etal-2020-nadi} which includes 21 labels.
The datasets were preprocessed by removing diacritics, URLs, and Twitter usernames while maintaining the same train, dev, and test splits for each dataset.
Moreover, we collated the tweets belonging to a particular user in the MADAR Twitter Corpus in groups of 5 before feeding them to the model. We refer to this preprocessed version as MADAR-Twitter-5 to avoid confusion with the publicly available original MADAR Twitter Corpus.

\paragraph{Experimental Setup}
Our models were fine-tuned for 10 epochs with a learning rate of 3e-5, batch size of 32, and a maximum sequence length of 128.
After the fine-tuning, we used the best checkpoints based on the dev sets to report results on the test sets using the macro $F_1$ score.
Moreover, for the MADAR-Twitter-5 evaluation, we took a voting approach. That is, each user in the dev and test sets is assigned to the most frequent predicted country label.
In case of a tie, we always pick the most frequent predicted country label based on the training set.

\subsection{Poetry Meter Classification}
\paragraph{Dataset}
We used the Arabic Poem Comprehensive Dataset (APCD)~\cite{yousef2019learning}, which is mostly in CA, to fine-tune our models to identify the meters of Arabic poems.
The dataset contains around 1.8M poems and covers 23 meters.
We preprocessed the dataset by removing diacritics from the poems and separated the halves of each verse by using the \texttt{[SEP]} token. We applied an 80/10/10 random split to create train, dev, and test sets respectively.

\paragraph{Experimental Setup}
We fine-tuned our models on a single GPU for 3 epochs with a learning rate of 3e-5, batch size of 32, and a maximum sequence length of 128.
After the fine-tuning, we used the best checkpoint based on the dev set to report results on the test set using the macro $F_1$ score.

\begin{table*}[ht!]
\setlength{\tabcolsep}{2pt}
\centering
\begin{tabular}{llc|ccccc;{1.5pt/1.5pt}c}
\toprule
\multirow{3}{*}{\textbf{Task}} & \multirow{3}{*}{\textbf{Subtask}} & \multirow{3}{*}{\textbf{Variant}} & \multicolumn{6}{c}{\textbf{\%Performance}} \\
&  &  & \textbf{MSA} & \textbf{MSA-1/2} & \textbf{MSA-1/4} & \textbf{MSA-1/8} &  \textbf{MSA-1/16} & \multirow{2}{*}{\textbf{Max-Min}} \\
&  &  & \textbf{(107GB)} & \textbf{(53GB)} & \textbf{(27GB)} & \textbf{(14GB)} &  \textbf{(6GB)} & \\
\hline
NER & ANERcorp & MSA & 82.4 & 82.0 & 82.1 & \textbf{82.6} & 80.8 & 1.9 \\
\hline
\multirow{3}{*}{POS} & PATB (MSA) & MSA & \textbf{98.3} & 98.2 & \textbf{98.3} & 98.2 & 98.2 & 0.1 \\
& ARZTB (EGY) & DA & 93.6 & 93.6 & \textbf{93.7} & 93.6 & 93.6 & 0.2 \\
& Gumar (GLF) & DA & \textbf{97.9} & \textbf{97.9} & \textbf{97.9} & \textbf{97.9} & \textbf{97.9} & 0.1 \\
\hline
\multirow{3}{*}{SA} & ASTD & MSA & \textbf{76.9} & 76.0 & 76.8 & 76.7 & 75.3 & 1.6 \\
& ArSAS & MSA & \textbf{93.0} & 92.6 & 92.5 & 92.5 & 92.3 & 0.8 \\
& SemEval & MSA & 72.1 & 70.7 & \textbf{72.8} & 71.6 & 71.2 & 2.0 \\
\hline
\multirow{4}{*}{DID} & MADAR-26 & DA & 62.6 & 62.0 & \textbf{62.8} & 62.0 & 62.2 & 0.8 \\
& MADAR-6 & DA & 91.9 & 91.8 & \textbf{92.2} & 92.1 & 92.0 & 0.4 \\
& MADAR-Twitter-5 & MSA & 77.6 & \textbf{78.5} & 77.3 & 77.7 & 76.2 & 2.3 \\
& NADI & DA & \textbf{24.9} & 24.6 & 24.6 & \textbf{24.9} & 23.8 & 1.1 \\
\hline
Poetry & APCD & CA & 79.7 & 79.9 & \textbf{80.0} & 79.7 & 79.8 & 0.3 \\
\hline
\hline
\multicolumn{2}{l}{\multirow{3}{*}{\textbf{Variant-wise-average}}}& MSA & \textbf{83.4} & 83.0 & 83.3 & 83.2 & 82.3 & 1.1 \\
& & DA & 74.2 & 74.0 & \textbf{74.3} & 74.1 & 73.9 & 0.4 \\
& & CA & 79.7 & 79.9 & \textbf{80.0} & 79.7 & 79.8 & 0.3 \\
\hdashline[1.5pt/1.5pt]
\multicolumn{3}{l|}{\textbf{Macro-average}} & \textbf{79.2} & 79.0 & \textbf{79.2} & 79.1 & 78.6 & 0.6 \\
\bottomrule
\end{tabular}
\caption{
\label{table:msa-result}
Performance of \camelbert~models trained on MSA datasets with different sizes.
We use the $F_1$ score as an evaluation metric for all tasks, except for POS tagging where we use accuracy.
Max-Min refers to the difference in performance among the models for each dataset.
Variant-wise-average refers to average over a group of tasks in the same language variant.
The best results among the models are in bold.
}
\end{table*}

\section{Evaluation Results and Discussion}
\label{sec:evaluation}
We first present an experiment where we investigate the effect of pre-training data size.
We then report on {\camelbert} models pre-trained on MSA, DA, and CA data, in addition to a model that is pre-trained on a mixture of these variants. 
We then provide a comparison against publicly available models.

\subsection{Models with Different Data Sizes}
\label{datasize-analysis}
To investigate the effect of pre-training data size on fine-tuning tasks, we pre-train MSA models in a controlled setting where we scale down the MSA pre-training size by a factor of two while keeping all other hyperparameters constant. We pre-train four {\camelbert} models on MSA data as follows:
MSA-1/2 (54GB, 6.3B words),
MSA-1/4 (27GB, 3.1B words),
MSA-1/8 (14GB, 1.5B words), and
MSA-1/16 (6GB, 636M words).
In Table \ref{table:msa-result}, we show the results on our fine-tuning subtasks.

We observe that the full MSA model and the MSA-1/4 model are on average the highest performing systems, even though the MSA-1/4 model was pre-trained on a quarter of the full MSA data.
The MSA-1/4 model wins or ties for a win in seven out of 12 subtasks, and it is also the best model on average in the DA and CA subtasks.

We also observe that different subtasks have different patterns.
For some subtasks, plateauing in performance happens rather early.
For instance, the performance on Gumar (GLF) does not change even if we increase the size.
Similarly, the difference in performance on PATB (MSA) is very small.
For other subtasks, the improvement is not consistent with the size, as seen in SemEval.
When we calculate the correlation between the performance and the pre-training data size, we note that ArSAS has a strong positive correlation of 0.96, however, MADAR-6 has a negative correlation of -0.62.
In fact, the average of the correlation of each of the 12 experiments is 0.25, which is not a strong pattern correlating size with performance.

These observations suggest that the size of pre-training data has limited and inconsistent effect on the fine-tuning performance.
This is consistent with \newcite{micheli-etal-2020-importance}, where they concluded that pre-training data size does not show a strong monotonic relationship with fine-tuning performance in their controlled experiments on French corpora.

\begin{table*}[!htbp]
\centering
\setlength{\tabcolsep}{3pt}
\begin{tabular}{llc|c;{1.5pt/1.5pt}c;{1.5pt/1.5pt}ccc;{1.5pt/1.5pt}c|c;{1.5pt/1.5pt}ccc}
\toprule
\multirow{2}{*}{\textbf{Task}} & \multirow{2}{*}{\textbf{Dataset}} & \multirow{2}{*}{\textbf{Variant}} & \multicolumn{6}{c}{\textbf{\%Performance}}    & \multicolumn{4}{c}{\textbf{\%OOV}} \\
& & & \textbf{Star} & \textbf{Mix}& \textbf{MSA} & \textbf{DA} & \textbf{CA} & \textbf{Max-Min} & \textbf{Mix} &\textbf{MSA} & \textbf{DA} & \textbf{CA}\\
 \hline
NER & ANERcorp & MSA & \textbf{82.4} & 80.8 & \textbf{\underline{82.4}} & 74.1 & 67.9 & 14.5 & 0.2 & \underline{0.2} & 1.4 & 4.2 \\
 \hline
\multirow{3}{*}{POS} & PATB (MSA) & MSA & \textbf{98.3} & 98.1 &  \textbf{\underline{98.3}} & 97.7 & 97.8 & 0.6 & 0.2 & \underline{0.2} & 0.9 & 3.0 \\
& ARZTB (EGY) & DA & \textbf{93.6} & \textbf{93.6} &  \textbf{\underline{93.6}} & 92.7 & 92.3 & 1.4 & 0.6 & \underline{0.8} & 1.0 & 7.3 \\
& Gumar (GLF) & DA & \textbf{98.1} & \textbf{98.1} & \underline{97.9} & \underline{97.9} & 97.7 & 0.2 & 0.2 & 0.8 & \underline{0.3} & 5.4 \\
 \hline
\multirow{3}{*}{SA} & ASTD & MSA & \textbf{76.9} & 76.3 & \textbf{\underline{76.9}} & 74.6 & 69.4 & 7.5 & 0.9 & \underline{1.1} & 1.2 & 5.3 \\
& ArSAS & MSA & \textbf{93.0} & 92.7 & \textbf{\underline{93.0}} & 91.8 & 89.4 & 3.6 & 1.3 & \underline{1.5} & 1.8 & 7.4 \\
& SemEval & MSA & \textbf{72.1} & 69.0 & \textbf{\underline{72.1}} & 68.4 & 58.5 & 13.6 & 1.9 & \underline{2.1} & 2.4 & 6.6 \\
 \hline
 \multirow{4}{*}{DID} & MADAR-26 & DA & \textbf{62.9} & \textbf{62.9} & \underline{62.6} & 61.8 & 61.9 & 0.8 & 0.4 & \underline{0.8} & \underline{0.8} & 7.5 \\
& MADAR-6 & DA & \textbf{92.5} & \textbf{92.5} & 91.9 & \underline{92.2} & 91.5 & 0.7 & 0.1 & 1.1 & \underline{0.2} & 8.1 \\
& MADAR-Twitter-5 & MSA & \textbf{77.6} & 75.7 & \textbf{\underline{77.6}} & 74.2 & 71.4 & 6.2 & 2.4 & \underline{2.6} & 3.0 & 6.7 \\
& NADI & DA & 24.7 & 24.7 & \textbf{\underline{24.9}} & 20.1 & 17.3 & 7.6 & 1.6 & \underline{2.0} & 2.4 & 8.1 \\
 \hline
Poetry & APCD & CA & \textbf{80.9} & 79.8 & 79.7 & 79.6 & \textbf{\underline{80.9}} & 1.3 & 0.4 & 1.1 & 2.7 & \underline{0.9} \\
 \hline
  \hline
\multicolumn{2}{l}{\multirow{3}{*}{\textbf{Variant-wise-average}}} & MSA & \textbf{83.4} & 82.1 & \textbf{\underline{83.4}} & 80.1 & 75.7 & 7.6 & 1.2 & \underline{1.3} & 1.8 & 5.5 \\
& & DA & \textbf{74.4} & \textbf{74.4} & \underline{74.2} & 72.9 & 72.1 & 2.1 & 0.6 & 1.1 & \underline{0.9} & 7.3 \\
& & CA & \textbf{80.9} & 79.8 & 79.7 & 79.6 & \textbf{\underline{80.9}} & 1.3 & 0.4 & 1.1 & 2.7 & \underline{0.9} \\
\hdashline[1.5pt/1.5pt]
\multicolumn{3}{l|}{\textbf{Macro-average}} & \textbf{79.4} & 78.7 & \underline{79.2} & 77.1 & 74.7 & 4.6 & 0.9 & \underline{1.2} & 1.5 & 5.9 \\
\bottomrule
\end{tabular}
\caption{\label{table:main}
Performance of \camelbert~models trained on MSA, DA, CA, and their Mix data.
Star refers to a way of choosing {\camelbert} models based on the language variant of the fine-tuning dataset.
We use the $F_1$ score as an evaluation metric for all tasks, except for POS tagging where we use accuracy.
Max-Min refers to the difference in performance among \camelbert's MSA, DA, and CA models only.
The best results among \camelbert's MSA, DA, and CA models are underlined.
The best results among \camelbert's MSA, DA, CA, Mix, and Star are in bold.
The OOV rate for each dataset is calculated based on the data used for pre-training each model.
We underline the lowest OOV value per dataset.
}
\end{table*}

\subsection{Models with Different Language Variants}
\label{subsec:model_variant}
Next, we explore the relationship between language variants in pre-training and fine-tuning datasets.

\subsubsection{MSA, DA, and CA}
\paragraph{Task Type Difference} We compare the behavior of three models pre-trained on MSA, DA, and CA data.
From Table \ref{table:main}, we observe that the difference in performance (Max-Min) among {\camelbert}'s MSA, DA, and CA models is 4.6\% on average, ranging from 0.2\% to 14.5\%.
To study trends by task type, we compute the average performance difference across the subtasks for each task.
NER is the most sensitive to the pre-trained model variant (14.5\%), followed by sentiment analysis (8.2\%), dialect identification (3.8\%), poetry classification (1.3\%), and POS tagging (0.7\%).
This indicates the importance of optimal pairing of pre-trained models and fine-tuning tasks.

On average the {\camelbertmsa} model performs best, and is the winner in 10 out of 12 subtasks.
The following are the two exceptions:
(a) the {\camelbertda} model performs best in the highly dialectal MADAR-6 subtask; and 
(b) the {\camelbertca} model outperforms other models in the poetry classification task, which is in classical Arabic.
These two exceptions suggest that performance in fine-tuning tasks may be associated with the variant proximity of the pre-training data to fine-tuning data; although we also acknowledge that {\camelbertmsa}'s data is two times the size of {\camelbertda}'s, and 18 times the size of {\camelbertca}'s, which may give {\camelbertmsa} an advantage.

\paragraph{OOV Effect} To further investigate the effect of variant proximity on performance, we compute the word out-of-vocabulary (OOV) rate of all fine-tuning test sets against the pre-training data, as a way to estimate their similarity.\footnote{We use a simple token as a unit, where we segment text with white space and punctuation.}
Note that {\camelbertmix}, where we concatenate MSA, DA, and CA pre-training data, has the lowest OOV rate by design.
In Table \ref{table:main}, we show the OOV rates for each dataset.

In all the cases, we obtain the best performance where the model has the lowest OOV rate.
To better understand the relationship between fine-tuning performance and OOV rates, we assessed the correlation between model performance and OOV rates for each dataset.
We found a strong negative correlation of -0.82 on average.
Interestingly, the {\camelbertca} model which was pre-trained only on 6 GB of data outperforms other models that are pre-trained on significantly larger data in the poetry classification task.
It is also worth mentioning that the {\camelbertca} model has the lowest OOV rate on the poetry dataset (0.9\%), while having access to approximately 18 times less data compared to the {{\camelbertmsa}} model (6GB vs 107GB).
This again suggests that the variant proximity of pre-training data to fine-tuning data is more important than the size of pre-training data.

\begin{table*}[ht]
\centering
\setlength{\tabcolsep}{1.5pt}
\begin{tabular}{llc|c;{1.5pt/1.5pt}cccc|cccccccc}
\toprule
\multirow{2}{*}{\textbf{Task}} & \multirow{2}{*}{\textbf{Dataset}} & \multirow{2}{*}{\textbf{Variant}} & \multicolumn{13}{c}{\textbf{\%Performance}} \\
 & & & \textbf{Star} & \textbf{Mix}& \textbf{MSA} & \textbf{DA} & \textbf{CA} & $X_1$ &  $X_2$ &  $X_3$ & $X_4$ & $X_5$ & $X_6$ & $X_7$ & $X_8$\\
 \hline
NER & ANERcorp & MSA & 82.4 & 80.8 & 82.4 & 74.1 & 67.9 & 76.7 & 82.8 & 82.0 & 80.3 & 77.3 & 82.0 & 79.3 & \textbf{83.6} \\
\hline
\multirow{3}{*}{POS} & PATB (MSA) & MSA & 98.3 & 98.1 & 98.3 & 97.7 & 97.8 & 97.9 & 98.2 & 98.3 & 98.3 & 98.1 & \textbf{98.4} & 98.0 & \textbf{98.4} \\
& ARZTB (EGY) & DA & 93.6 & 93.6 & 93.6 & 92.7 & 92.3 & 92.0 & 93.0 & \textbf{94.1} & 93.1 & 93.3 & 93.6 & 93.5 & 93.6 \\
& Gumar (GLF) & DA & \textbf{98.1} & \textbf{98.1} & 97.9 & 97.9 & 97.7 & 97.4 & 97.8 & \textbf{98.1} & 97.8 & 97.7 & 98.0 & 97.9 & 97.9 \\
\hline
\multirow{3}{*}{SA} & ASTD & MSA & 76.9 & 76.3 & 76.9 & 74.6 & 69.4 & 64.5 & 74.2 & \textbf{78.1} & 73.5 & 74.2 & 74.3 & 77.0 & 74.9 \\
& ArSAS & MSA & 93.0 & 92.7 & 93.0 & 91.8 & 89.4 & 88.4 & 91.5 & \textbf{93.3} & 92.3 & 91.3 & 92.2 & 92.9 & 91.9 \\
& SemEval & MSA & 72.1 & 69.0 & 72.1 & 68.4 & 58.5 & 57.5 & 69.5 & \textbf{72.7} & 69.5 & 70.7 & 70.0 & 70.4 & 70.2 \\
\hline
\multirow{4}{*}{DID} & MADAR-26 & DA & \textbf{62.9} & \textbf{62.9} & 62.6 & 61.8 & 61.9 & 60.4 & 61.9 & 62.2 & 58.4 & 59.8 & 59.1 & 61.2 & 60.7 \\
& MADAR-6 & DA & \textbf{92.5} & \textbf{92.5} & 91.9 & 92.2 & 91.5 & 90.8 & 91.9 & 92.3 & 90.8 & 91.5 & 91.4 & 92.1 & 91.4 \\
& MADAR-Twitter-5 & MSA & 77.6 & 75.7 & 77.6 & 74.2 & 71.4 & 71.8 & \textbf{79.0} & \textbf{79.0} & 74.7 & 77.7 & 77.6 & 78.6 & 76.5 \\
& NADI & DA & 24.7 & 24.7 & 24.9 & 20.1 & 17.3 & 16.7 & 21.1 & 24.5 & 24.0 & 25.0 & 21.3 & \textbf{27.0} & 24.6 \\
\hline
Poetry & APCD & CA & \textbf{80.9} & 79.8 & 79.7 & 79.6 & \textbf{80.9} & 78.8 & 79.6 & 79.9 & 78.8 & 79.1 & 79.1 & 79.0 & 78.4 \\
\hline
\hline
 \multicolumn{2}{l}{\multirow{3}{*}{\textbf{Variant-wise-average}}} & MSA & 83.4 & 82.1 & 83.4 & 80.1 & 75.7 & 76.1 & 82.5 & \textbf{83.9} & 81.4 & 81.6 & 82.4 & 82.7 & 82.6 \\
& & DA & \textbf{74.4} & \textbf{74.4} & 74.2 & 72.9 & 72.1 & 71.5 & 73.1 & 74.2 & 72.8 & 73.5 & 72.7 & 74.3 & 73.6 \\
& & CA & \textbf{80.9} & 79.8 & 79.7 & 79.6 & \textbf{80.9} & 78.8 & 79.6 & 79.9 & 78.8 & 79.1 & 79.1 & 79.0 & 78.4 \\
\hdashline[1.5pt/1.5pt]
\multicolumn{3}{l|}{\textbf{Macro-average}} & 79.4 & 78.7 & 79.2 & 77.1 & 74.7 & 74.4 & 78.4 & \textbf{79.5} & 77.6 & 78.0 & 78.1 & 78.9 & 78.5 \\
\bottomrule
\end{tabular}
\caption{\label{table:all_models}
Performance of \camelbert~models and other existing models.
We use the $F_1$ score as an evaluation metric for all tasks, except for POS tagging where we use accuracy.
Star refers to a way of choosing \camelbert~models based on the language variant of the fine-tuning dataset.
$X_1, \cdots, X_8$ corresponds to the models in Table \ref{table:model_comparison}.
The best results among the models are in bold.
}
\end{table*}

\subsubsection{Mix of MSA, DA, and CA}
To further study the interplay of language variants and pre-training data size, we pre-trained a model 
({\camelbertmix})
on the concatenation of the MSA, DA, and CA datasets. 
This is the largest dataset used to pre-train an Arabic language model to date.
As shown in Table \ref{table:main}, the {\camelbertmix} model improves over other models in three cases, all of which are dialectal, suggesting that the {\camelbertmix} model does better in some dialectal context.
However, we do not see an increase in performance in other cases when compared with the best performing model, although the size of the pre-training data and the variety of the data are increased.
This suggests that having a wide language variety in pre-training data can be beneficial for DA subtasks, whereas variant proximity of pre-training data to fine-tuning data is important MSA and CA subtasks.

\subsubsection{Selecting an Optimal Model}
Taking these insights into consideration, one cannot help but consider the exciting possibility of a system-selection ensembling approach that can help users make decisions with reasonable expectations using what they know of their specific tasks.
We outline here such a setup: the user has access to three versions of the models: {\camelbert}'s CA, MSA, and Mix.
If the task data is known a priori to be CA, then we select the {\camelbertca} model;
if the task data is known to be MSA, we select the {\camelbertmsa} model;
otherwise, we use the {\camelbertmix} model (for dialects, i.e.).
We report on this model in Table \ref{table:main} and \ref{table:all_models} as {\camelbertstar}.

It is noteworthy that this model is not the same as oracularly selecting the best performer among our four models (MSA, DA, CA, and Mix).
In fact, it is lower in performance than such oracular system as the {\camelbertmsa} model performs better than {\camelbertmix} model in  NADI.
We do not claim here that this is a foolproof method; however, it is an interesting candidate for {\it common wisdom} of the kind we are hoping to develop through this effort.

\subsection{Comparison with Existing Models}
Table~\ref{table:all_models} compares our work with other existing models.
We do not use models that require morphological pre-tokenization to allow direct comparison, and also because existing tokenization systems are mostly focused on MSA or EGY \cite{Pasha:2014:madamira,Abdelali:2016:farasa,obeid-etal-2020-camel}. 

We are aware that design decisions such as vocabulary size and number of training steps are not the same across these eight existing pre-trained models, which might be a contributing factor to their varying performances. We plan to investigate the effects of such decisions in future work.

\paragraph{Task Performance Complementarity} The best model on average is AraBERTv02 ($X_3$); it wins or ties for a win in six out of 12 subtasks (four MSA and two DA).
Our {\camelbertstar} is second overall on average, and it wins or ties for a win in four out of 12 subtasks (three DA, one CA).
Interestingly, the two systems are complementary in their performance and between the two they win or tie for a win in nine out of 12 subtasks.
The three remaining subtasks are won by MARBERT ($X_7$) (NADI, DA), ARBERT ($X_8$) (ANERcorp, MSA; PATB, MSA), and GigaBERT ($X_6$) (PATB, MSA).
In practice, such complementarity can be exploited by system developers to achieve higher overall performance.

\paragraph{Size and Performance}
Considering the data size and performance of the other pre-trained models ($X_1$ to $X_8$), we observe a similar trend to our {\camelbert} models.
AraBERTv02 ($X_3$) is the best on average, with only 77GB of pre-training data.
AraBERTv01 ($X_2$) is the smallest (24GB); however, on average it outperforms other models pre-trained on much larger datasets, such as and ArabicBERT ($X_4$, 95GB) and multi-dialectal Arabic BERT ($X_5$, 95GB with 10M tweets).
This confirms that pre-training data size may not be an important factor to fine-tuning performance, as we showed in Section~\ref{datasize-analysis}.

\paragraph{Variant Proximity and Performance}
When we examine the proximity in terms of language variants of the pre-training data and the fine-tuning data across the eight existing pre-trained models, we observe the following.
First, the monolingual MSA models ($X_2$, $X_3$, $X_4$, $X_8$) are better performers than the mixed models ($X_5$, $X_7$) on average (78.5\% and 78.4\%, respectively).\footnote{The average over macro-average performances.}
Second, the monolingual MSA models perform better than the mixed models in MSA subtasks on average (82.6\% and 82.1\%, respectively), while the mixed models perform better than the MSA models in DA subtasks on average (73.5\% and 73.9\%, respectively).\footnote{The average over variant-wise-average performances.}
This result is consistent with our analysis of the {\camelbertmix} and the {\camelbertmsa} models in Section~\ref{subsec:model_variant}, where we found that the {\camelbertmix} model is the best choice for DA subtasks, whereas the {\camelbertmsa} model is the best in MSA subtasks.


\paragraph{\it On MARBERT and ARBERT} In another study that compared models pre-trained on MSA alone or a mix of MSA and DA data, \newcite{abdulmageed2020arbert} reported that
MARBERT ($X_7$, pre-trained on  MSA-DA mix) is more powerful than ARBERT ($X_8$, pre-trained on MSA).
In our study, we do replicate their specific relative performance in terms of macro-average in our experiments (78.9\% for MARBERT and 78.5\% for ARBERT).
It is not clear why MARBERT and ARBERT do not exhibit similar trends as observed in the analysis of our own {\camelbert} models and other existing models. This may be attributed to numerous factors such as the degree of MSA-DA mixture, genre, and the pre-training procedure details.
It is also worth noting that the data used to pre-train our {\camelbertmsa} model is a subset of the data used to pre-train our {\camelbertmix} model, whereas the pre-training data for MARBERT and ARBERT are derived from different data sources.

\section{Conclusion and Future Work}
\label{sec:conclusion}
In this paper, we investigated the interplay of size, language variant, and fine-tuning task type
in Arabic pre-trained language models using carefully controlled experiments on a number of Arabic NLP tasks.
Our results show that pre-training data and subtask data variant proximity is more important than pre-training data size.
We confirm these results on existing models.
We exploit this insight in defining an optimized system selection model for the studied tasks. 
We make all of our created models and fine-tuning code publicly available.

In future work, we plan to explore other design decisions that may contribute to the fine-tuning performance, including vocabulary size, tokenization techniques, and additional data mixtures. We also plan to utilize {\camelbert} models in a number of other Arabic NLP tasks, and integrate them in the open-source toolkit, CAMeL~Tools \cite{obeid-etal-2020-camel}.

\section*{Acknowledgment}
This research was supported with Cloud TPUs from Google’s TensorFlow Research Cloud (TFRC).
This work was also carried out on the High Performance Computing resources at New York University Abu Dhabi.
The first and second authors were supported by the New York University Abu Dhabi
Global PhD Student Fellowship program.
We thank Salam Khalifa, and Ossama Obeid for helpful discussions.
We also thank the anonymous reviewers for their valuable comments.
\bibliography{anthology, camel-bib-v2, extra}
\bibliographystyle{acl_natbib}

\appendix

\end{document}